\documentclass[conference]{IEEEtran}
\IEEEoverridecommandlockouts
\usepackage{cite}
\usepackage{amsmath,amssymb,amsfonts}
\usepackage{algorithmic}
\usepackage{graphicx}
\usepackage{textcomp}
\usepackage{xcolor}
\usepackage[whole]{bxcjkjatype}
\def\BibTeX{{\rm B\kern-.05em{\sc i\kern-.025em b}\kern-.08em
    T\kern-.1667em\lower.7ex\hbox{E}\kern-.125em}}
\begin{document}

\title{What is the Reward for Handwriting? ---\\ Handwriting Generation \\ by Imitation Learning}

% \author{Anonymous}
\author{\IEEEauthorblockN{Keisuke Kanda, Brian Kenji Iwana, and Seiichi Uchida}
\IEEEauthorblockA{\textit{Department of Advanced Information Technology} \\
\textit{Kyushu University}\\
Fukuoka, Japan \\
keisuke.kanda@human.ait.kyushu-u.ac.jp, \{iwana,uchida\}@ait.kyushu-u.ac.jp}
}

\maketitle

\begin{abstract}
Analyzing the handwriting generation process is an important issue and has been tackled by various generation models, such as kinematics based models and stochastic models.
In this study, we use a reinforcement learning (RL) framework to realize handwriting generation with the careful future planning ability. In fact, the handwriting process of human beings is also supported by their future planning ability; for example, the ability is necessary to generate a closed trajectory like `0' because any shortsighted model, such as a Markovian model, cannot generate it. For the algorithm, we employ generative adversarial imitation learning (GAIL). Typical RL algorithms require the manual definition of the reward function, which is very crucial to control the generation process. In contrast, GAIL trains the reward function along with the other modules of the framework. In other words, through GAIL, we can understand the reward of the handwriting generation process from handwriting examples. Our experimental results qualitatively and quantitatively show that the learned reward catches the trends in handwriting generation and thus GAIL is well suited for the acquisition of handwriting behavior.
\end{abstract}


%%%%%%%%%%%%%%%%%%%%%%%%%%%%%%%%%%%%%%%%%%%%%%%%%%%%
\section{Introduction\label{sec:introduction}}
%%%%%%%%%%%%%%%%%%%%%%%%%%%%%%%%%%%%%%%%%%%%%%%%%%%%
The generation process of handwriting trajectories is an important research target. For example, an accurate generative model is required for more accurate handwriting recognition. Another example is that the medical diagnosis of Parkinson's, Alzheimer's, and dyslexia often depends on the analysis of patients' handwriting trajectories.
Modeling of the individuality in handwriting trajectories is also important for forensic applications. A more scientific role of the generative model is to understand, ``what is handwriting.'' In other words, among temporal trajectories on two-dimensional space, which factors discriminate handwriting trajectories from other trajectories. \par
Various generative models of handwriting have been proposed so far. Machine learning methods, or stochastic modeling methods, are often used for recognition tasks.
Hidden Markov models and recurrent neural networks, such as long short-term memory (LSTM)~\cite{Hochreiter_1997} and gated recurrent unit (GRU)~\cite{cho2014learning}, are typical choices. In contrast to those bottom-up (i.e., data-driven) models, physical models are also proposed. One of the most famous models is the delta-lognormal model, which is based on human kinematic theory~\cite{plamondon}. Reaching models, such as the minimum jerk model~\cite{flash-hogan}, are also useful to understand the physical constraint in handwriting generation. Those physical models are top-down models and designed by some assumption of human arm movements. Another type of the handwriting generation model is a classical but interesting work by Kellogg~\cite{kellogg}; she observed a million of Children's drawings and found a rule (i.e., a model) of the process that children acquire handwriting skill from twenty basic scribbles to hand-drawn pictures (early pictorialism). \par
%
%=========================
\begin{figure}
\centering
\includegraphics[width=0.925\linewidth]{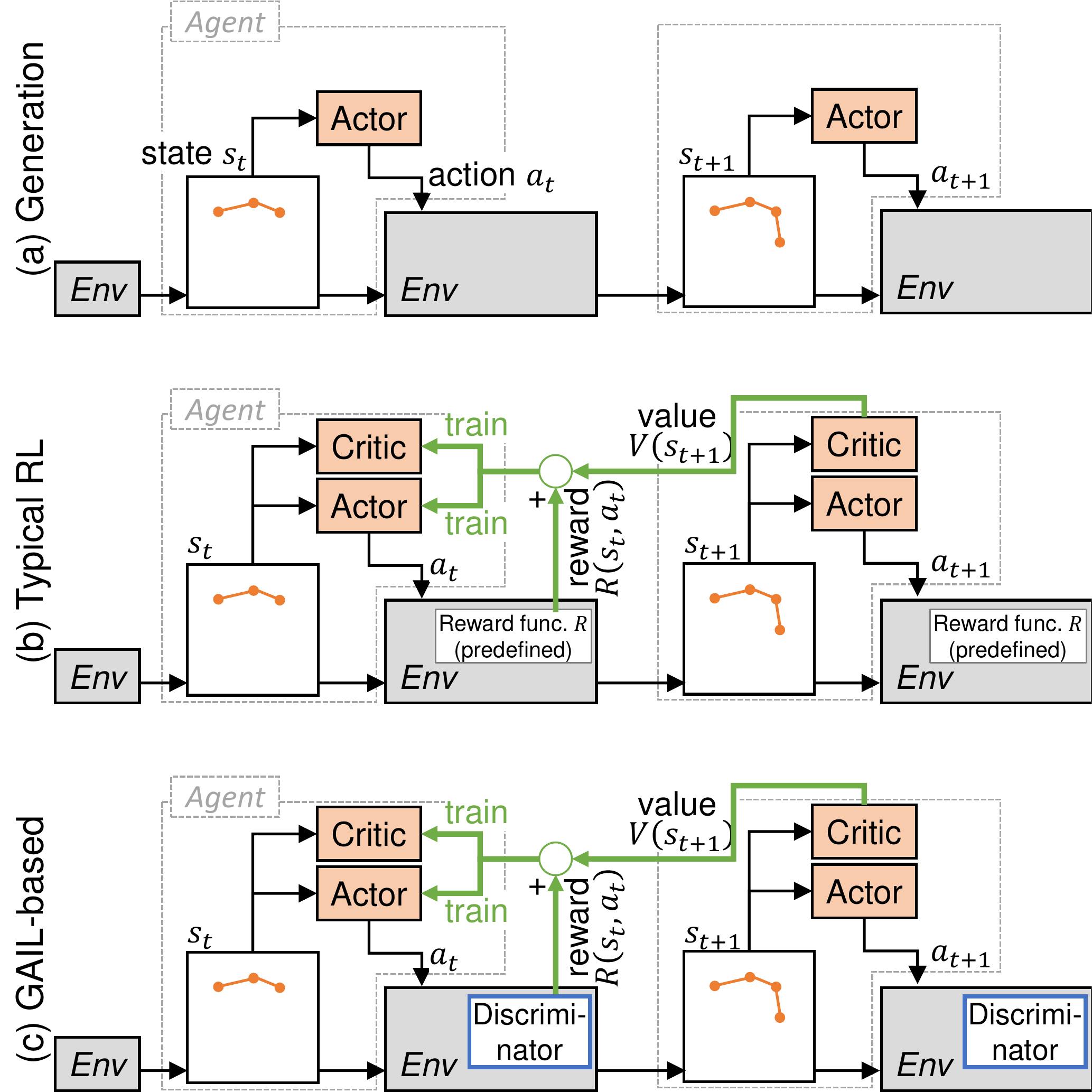}
\caption{(a)~Handwriting generation by a model trained by RL. (b)~Training procedure of typical RL. (c)~Training procedure of our GAIL-based model.\vspace{-3mm}}
\label{fig:rl}
\end{figure}
%=========================

Considering the actual process of acquiring our handwriting generation skills, reinforcement learning (RL) can be a choice to formulate the generation model. Fig.~\ref{fig:rl}~(a) illustrates the model. The {\em actor} is the writer and the {\em action} is the pen-tip movement at $t$. By making actions from $t=1$ to $T$, a handwriting trajectory is generated as a sequence of {\em states}; the state at $t$ corresponds to a partial trajectory until $t$.\par
In general, the training process of RL mimics the skill acquisition process via trial-and-error between an agent and the environment. Fig.~\ref{fig:rl}~(b) shows the typical training procedure, called the actor-critic algorithm. The actor in the agent is trained to make an action to get a higher {\em reward} from the {\em environment}. The reward is the evaluation of the action at $t$, that is, the ``goodness'' of the pen-movement.  For handwriting, the environment determines the current state and gives visual feedback to the writer as the reward.\par
The highlight of using RL for modeling the handwriting generation process is that the actor is trained by not only instantaneous feedback (i.e., the reward at $t$) but also a careful future evaluation, called {\em value}. The value at $t$ is an estimation of the total reward after $t$\footnote{Using the discount technique, the total reward converges to a certain value.}. As shown in Fig.~\ref{fig:rl}~(b), the {\em critic} gives the value. Our writing process is also not just instantaneous. For generating a handwriting trajectory whose entire shape has a good balance, we always make the careful future planning of the pen-tip movement. For example, when we write `0', we move the pen-tip with a future plan to make a closed curve.
\par
One hurdle on using RL is how to define the reward function. In the typical framework of Fig.~\ref{fig:rl}~(b), the reward function is manually predefined by the user. However, we do not know what is the appropriate reward function for handwriting generation. The definition of the reward function is very crucial for all the other functions (i.e., the actor and the critic), and therefore any intuitive definition must be avoided.\par
The purpose of this paper is to model the handwriting generation process through imitation learning~\cite{Hussein_2017}, which is a version of RL, and acquire handwriting behavior. In imitation learning, the reward function is also trained along with other functions, by referring to expert knowledge (which corresponds to real handwriting trajectories in our case). Accordingly, we can avoid the intuitive definition of the reward function. Moreover, by observing the learned reward, we can understand the underlying reward for handwriting.\par
As for the algorithm of imitation learning, we use a modified Generative Adversarial Imitation Learning (GAIL)~\cite{GAIL} model. Fig.~\ref{fig:rl}~(c) shows our GAIL-based training for the handwriting generation model. GAIL is an imitation learning framework based on Generative Adversarial Networks (GAN)~\cite{goodfellow2014generative}. GAN is well-known for its alternating training of a generator and a discriminator. In GAIL, the generator corresponds to the actor and the discriminator corresponds to the reward function. Through their adversarial training, the discriminator is trained to evaluate how well the generated trajectory looks like handwriting and gradually acquires the appropriate evaluation skill as the reward function.\par
The main contributions of this paper are summarized as follows:
\begin{itemize}
     \item To the authors' best knowledge, this is the first trial to use an imitation learning framework for modeling the handwriting generation process.
     \item  The observation of the learned rewards, as well as the generated trajectories, shows that our GAIL-based model learns the trends in the handwriting process.
     \item A quantitative evaluation shows that the learned model can generate handwriting samples more accurately than a CNN-based generation model. This proves that the future planning ability of our model is more appropriate to model the handwriting generation process.
\end{itemize}

%%%%%%%%%%%%%%%%%%%%%%%%%%%%%%%%%%%%%%%%%%%%%%%%%%%%
\section{Related work}
%%%%%%%%%%%%%%%%%%%%%%%%%%%%%%%%%%%%%%%%%%%%%%%%%%%%
\subsection{Imitation learning}
The general idea of teaching models through imitation has a rich history~\cite{Hussein_2017}. However, recently, it has gained more attention due to the need to solve complex tasks with minimal expert knowledge. Specifically, imitation learning is the problem setting of learning to perform tasks through the behavior of a given expert.
In this setting, the learning takes place by rewarding the model based on how well it mimics the expert rather than direct knowledge or feedback about the task. \par
One simple approach is behavior cloning~\cite{Pomerleau_1991}.
In a seminal work, Pomerleau~\cite{pomerleau1989alvinn} used behavior cloning to train a neural network for autonomous driving.
This is done by using a state-action pair consisting of a current state and the action of the expert.
Through this, supervised learning is performed.
For example, in the case of~\cite{pomerleau1989alvinn,Pomerleau_1991}, a view of the road was provided to an Autonomous Land Vehicle In a Neural Network~(ALVINN) with the output prediction consisting of a direction to travel.
More recently, behavior cloning has also been used for robotics~\cite{Niekum_2014} and aerial vehicles~\cite{Morales_2004,Giusti_2016}. \par
However, while behavior cloning has been effective classically, it suffers from the problem of compounding errors~\cite{ross2010efficient}, dataset bias~\cite{Codevilla_2019}, and generalization issues~\cite{Codevilla_2019}.
This is because behavior cloning is a Markovian decision process that is heavily dependent on the data.
A faulty state or unknown situation can lead to errors in the subsequent states.
In order to overcome this, Inverse Reinforcement Learning (IRL) methods were introduced~\cite{Russell_1998}.
Instead of predicting the action as in behavior cloning, IRL methods aim to predict the reward function from the expert's demonstrations.
By predicting the reward function, the agent is able to overcome the single-step errors in decisions of simple behavior cloning.
However, many IRL methods have problems such as computation time and issues learning the reward function~\cite{finn2016guided,GAIL}.
In addition, IRL algorithms only learn the reward function from the expert and do not indicate the required action directly.

%------------------------------------------------
\subsection{Reinforcement learning for handwriting}
In spite of RL's promising properties, there are few handwriting generation models based on it.
In one example, Wada and Sumita~\cite{wada} used RL to determine anchor points (via-points) for handwriting.
Their reward function is explicitly designed as the sum of an external reward and an internal reward.
The former is the similarity to a reference sample and the latter is to evaluate the torque around the writer's shoulder and elbow.
Muhammad et al.~\cite{muhammad} formulate the stroke removal process from handwritten sketches in an RL framework.
Their reward function is also explicitly designed by a classifier to evaluate the recognizability of the sketch after the stroke removal. %\par
Chao et al.~\cite{chao} proposed an arm robot that can draw Chinese calligraphy. Recently, Wu et al.~\cite{wu} have proposed its extended version. Their models are based on GAN and thus very similar to GAIL; however, their purpose is rather to realize a practical robotic system and not focused on the analysis of the learned rewards to understand the trends and characteristics of handwriting trajectories.\par
%
% Doya~\cite{doya} proposed a reinforcement learning model of basal ganglia,
% based on the fact that dopamine is given as a reward to the ganglia neurons  after the successful trial of a specific reaching task (i.e., a constrained arm motion task.)
%
In addition to handwriting generation, there are several applications of RL on handwriting text recognition. For example, in a recent work~\cite{Qiao_2018}, hidden features from a deep belief network are extracted and used as the state and the reward determines the class of the character.
In another example, Gui et al.~\cite{gui2018} proposed the use of RL, called a policy network, to select regions for the next inference.

%%%%%%%%%%%%%%%%%%%%%%%%%%%%%%%%%%%%%%%%%%%%%%%%%%%%
\section{Interpretation of handwriting generation in the terminology of reinforcement learning\label{sec:RL-for-HW}}
%%%%%%%%%%%%%%%%%%%%%%%%%%%%%%%%%%%%%%%%%%%%%%%%%%%%
In this section, we detail how the handwriting generation process is described in the terminology of RL, although it was already outlined in Section~\ref{sec:introduction}. Hereafter, we assume the trajectory is drawn on the $[0,1]\times [0,1]$ plane.\par
Each {\em state} $s_t$ corresponds to a handwriting trajectory with length $t\in \{1, 2,\ldots, T\}$.
We used a fixed-dimensional representation for each state by zero-padding. Specifically, each state is represented as a fixed-length sequence $(x_1,y_1,l_1),\ldots,\allowbreak (x_t,y_t,l_t),\ldots,\allowbreak  (x_T,y_T,l_T)$,
where $(x_t, y_t)\in [0,1]^2$ represents the pen-tip position at $t$. For $t'>t$, $(x_{t'},y_{t'})=(0,0)$, i.e., zero-padded. The variable $l_t$ indicates the zero-pad part; that is, $l_{t'}=1$ for $t'\leq t$ and $l_{t'}=0$ for $t'>t$. \par
The {\em action} $a_t$ corresponds to the next pen-tip position $(x_{t+1}, y_{t+1})$.  In our model, no hard constraint, such as continuity, is imposed on the action; therefore, theoretically, $(x_{t+1}, y_{t+1})$ can be very distant from $(x_t, y_t)$.\par
By the action $a_t$, the {\em environment} will make the state transition from $s_t$ to $s_{t+1}$.
In our model, the environment is the paper that gives visual feedback to the writer. For example, if $a_t=(10,20)$, the environment gives the new state $s_{t+1}$ just by replacing $(x_{t+1},y_{t+1},l_{t+1})=(0,0,0)$ of $s_t$ with $(10,20,1)$; on the paper, the trajectory becomes a bit longer with a new point $(10,20)$. \par
The {\em reward} $R(s_t, a_t)$ and the value play the most important role in the RL framework.
The reward evaluates the goodness of the action $a_t$ for the current state $s_t$. In our model, roughly speaking, the reward evaluates the goodness of the handwriting trajectory represented as the new state $s_{t+1}$. \par
The {\em value} $V(s_t)$ is an estimation of the total reward {\em after} $t$. This means the value evaluates how good $s_t$ is by considering huge possibilities of the succeeding actions (equivalently, states) after $s_t$. As noted in Section~\ref{sec:introduction}, our writing process is not just Markovian. It is natural to assume that we make a next pen movement by careful future planning with the help of visual feedback from the trajectory in the middle of writing on the paper.
If this assumption is valid, the RL framework is very suitable for modeling the handwriting generation process.
\par
%=========================
\begin{figure}[t]
\centering
\includegraphics[width=0.95\linewidth]{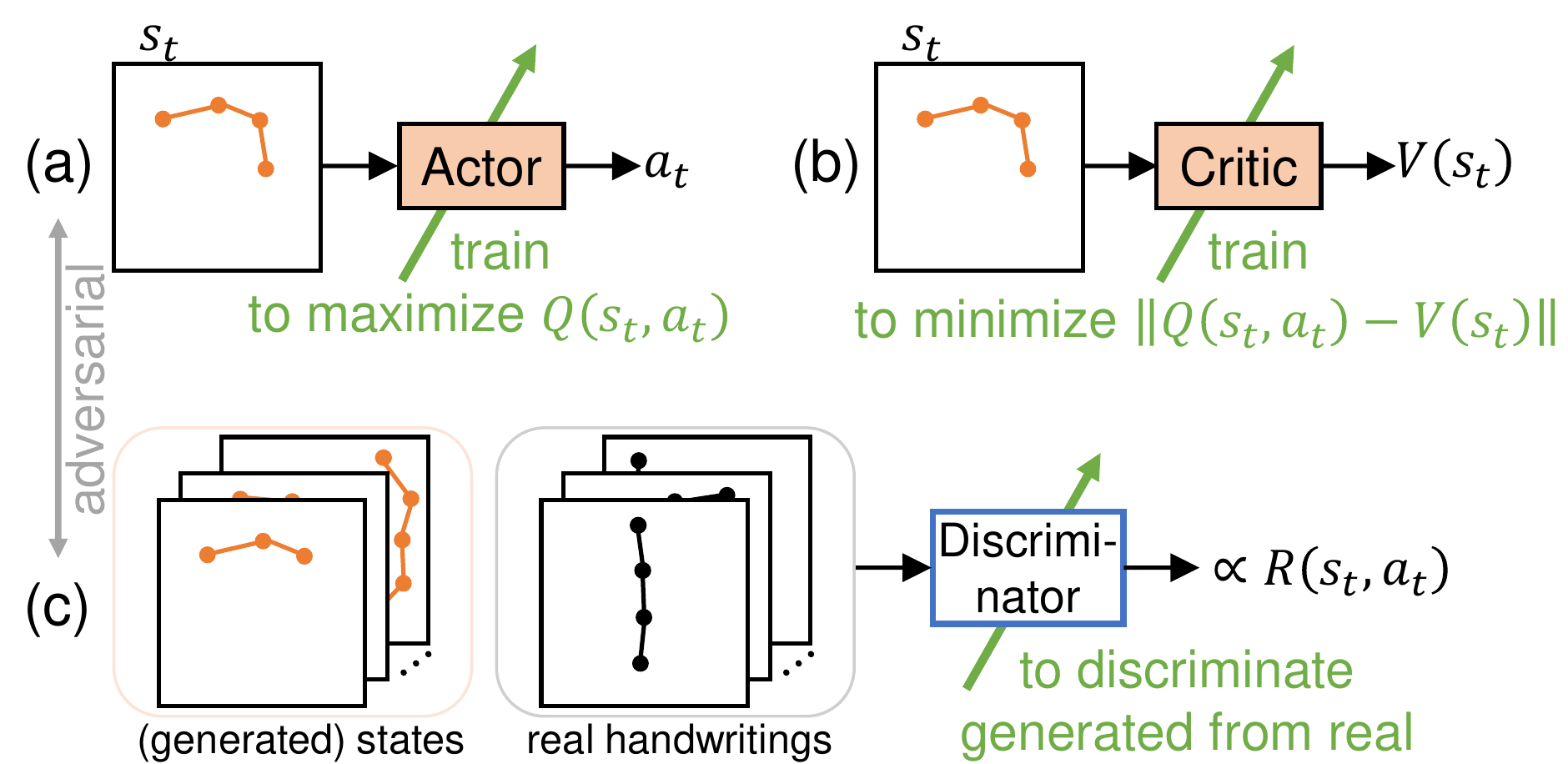}\\[-5mm]
\caption{Training (a)~the actor, (b)~the critic, and (c)~the discriminator.}
\label{fig:train}
\end{figure}
%=========================

%=========================
\begin{figure*}
\begin{flushleft}
\includegraphics[width=\linewidth]{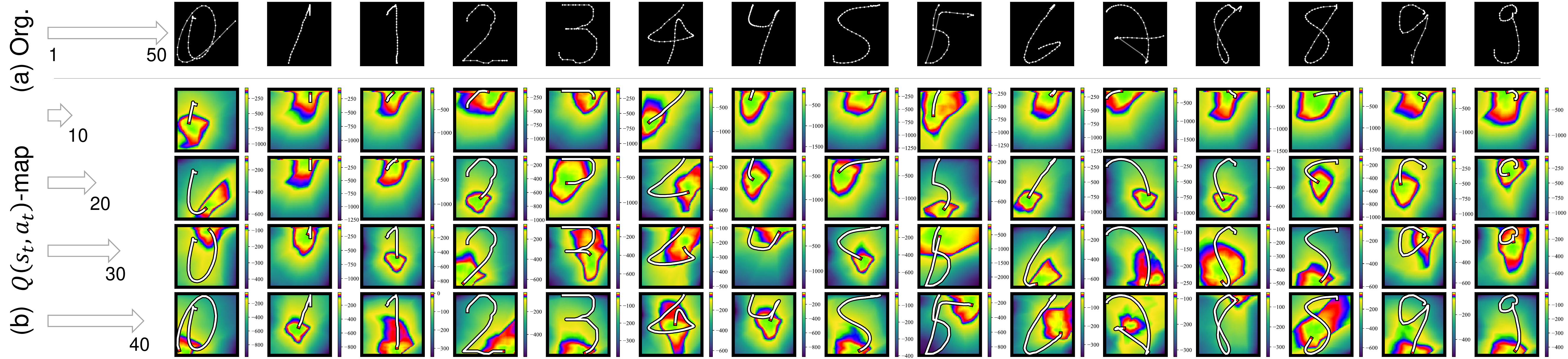}\vspace{-5mm}
\end{flushleft}
\centering
\caption{(a)~Original trajectory. (b)~Visualization of the learned values as $Q(s_t,a_t)$-map at $t=10, 20, 30$, and $40$.}
\label{fig:learned-result}

\begin{flushleft}
\includegraphics[width=0.985\linewidth]{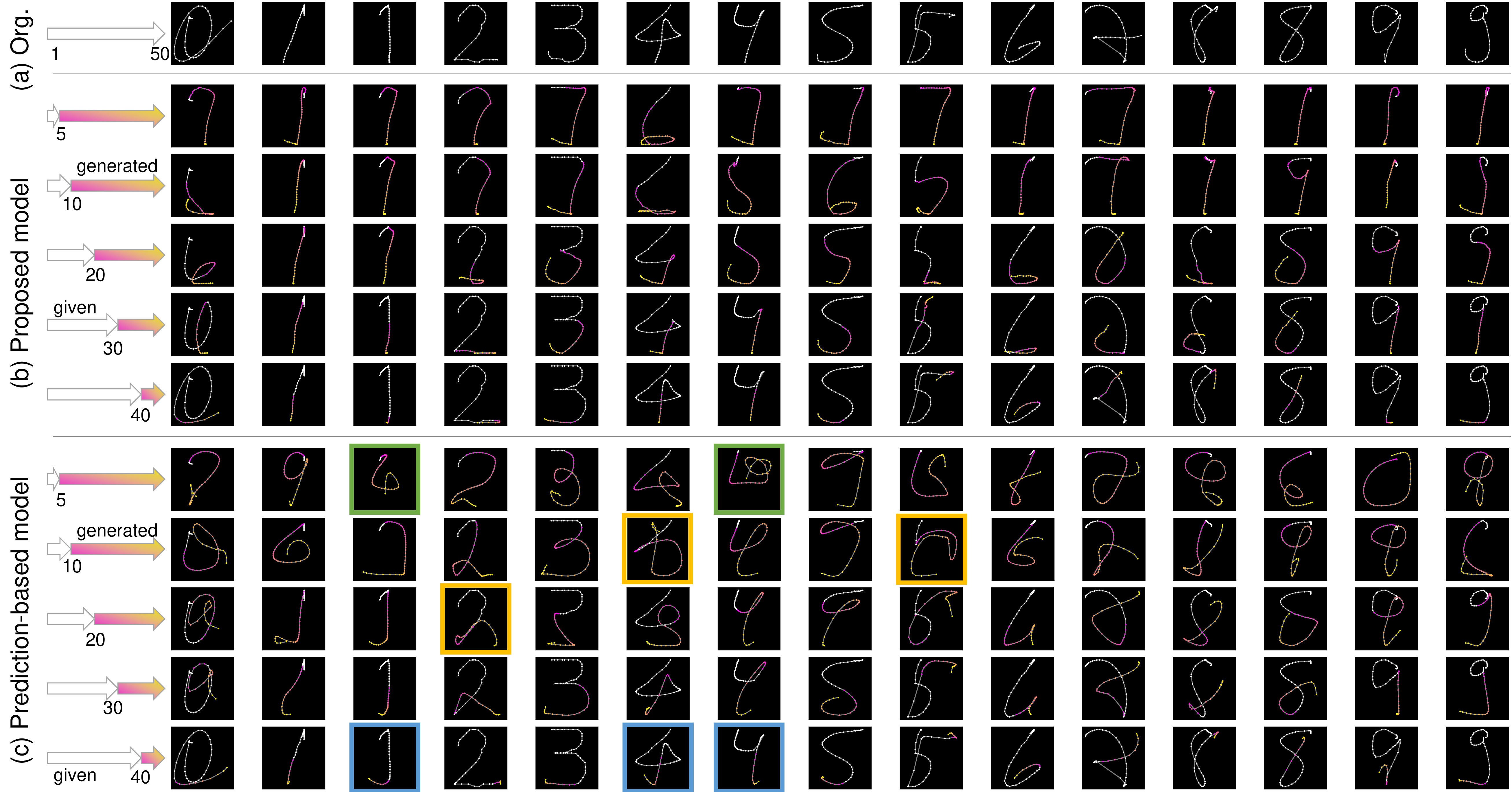}\vspace{-5mm}
\end{flushleft}
\centering
\caption{(a)~Original trajectory.
(b)~Generated trajectory by our GAIL-based model with the beginning part of (a). (c)~Generated trajectory by the prediction-based model. Note that this is not the task of predicting the original trajectory.}
\label{fig:generated}
\end{figure*}
%=========================

%%%%%%%%%%%%%%%%%%%%%%%%%%%%%%%%%%%%%%%%%%%%%%%%%%%%
\section{Handwriting generation model based on GAIL\label{sec:GAIL}}
%%%%%%%%%%%%%%%%%%%%%%%%%%%%%%%%%%%%%%%%%%%%%%%%%%%%
As shown in Fig.~\ref{fig:rl}~(c), GAIL~\cite{GAIL} has
three main modules, i.e., the actor, the critic, and the discriminator. All of them are implemented as a 1D-convolutional neural network (1D-CNN), whose details will be explained in Section~\ref{sec:modules}. The {\em actor} generates the action $a_t$ for the current state $s_t$. The {\em critic} evaluates the value $V(s_t)$ of the current state $s_t$. The {\em discriminator} gives the reward $R(s_t, a_t)$. Hereafter, we outline their properties; for the further detail of GAIL, please refer to its original literature~\cite{GAIL}, although we have made a modification to it as noted later.\par
Fig.~\ref{fig:train} illustrates the training procedure of
those modules. As shown in (a), the actor is trained to maximize $Q(s_t, a_t)$, which is defined as
$$
Q(s_t, a_t) = R(s_t, a_t) + \gamma V(s_{t+1}).
$$
Roughly speaking, $Q(s_t, a_t)$  evaluates the goodness of $a_t$ {\em while considering not only the instantaneous reward $R(s_t, a_t)$ but also its future rewards.}  This means that the trained actor has a future planning ability and thus provides the action $a_t$ that maximizes future rewards. \par
As shown in Fig.~\ref{fig:train}~(b), the critic is trained to provide $Q(s_t,a_t)$ that minimizes the L2 loss between $Q(s_t,a_t)$ and $V(s_t)$, where $a_t$ is given by the actor. This minimization relies on the fact that the ideal $V(s_t)$ should satisfy the following equation:
$$
V(s_t) = R(s_t, a_t) + \gamma V(s_{t+1}).
$$
The actor and the critic are trained simultaneously by Deep Deterministic Policy Gradient (DDPG)~\cite{DDPG}.
\par
As noted in Section~\ref{sec:introduction}, one of the most promising properties of GAIL is that it has the ability to train the reward function as the discriminator. Fig.~\ref{fig:train}~(c) shows the training process of the discriminator. Like the orthodox GAN's discriminator, it is trained to discriminate the generated trajectories (represented as $s_t$) from real handwriting trajectories. Since the actor generates the trajectories, this is the adversarial training between the actor and the discriminator.
After the training, the discriminator output can be used to evaluate how much the generated trajectory looks like a real handwriting trajectory \footnote{Precisely speaking, the output $\in [0,1]$ of the discriminator is fed to the logit function to have the reward $\in (-\infty, \infty)$.}.\par
Our implementation of GAIL is a modified version of the original GAIL~\cite{GAIL}. The original GAIL follows ``model-free'' formulation, where the environment is a black-box; that is, the derivation of $s_{t+1}$ from $s_t$ and $a_t$ is not explicitly formulated as a function. In contrast, the environment of our task
is very simple and explicit; $s_{t+1}$ is provided just by putting $a_t$ into $s_t$, as noted in Section~\ref{sec:RL-for-HW}. This means our GAIL results in the simpler ``model-based'' formulation.
Specifically, the critic needs to learn $Q(s_t, a_t)$ in the original GAIL and rather than the simpler $V(s_t)$ in ours; this is beneficial to simplify the network structure of the critic.

%%%%%%%%%%%%%%%%%%%%%%%%%%%%%%%%%%%%%%%%%%%%%%%%%%%%
\section{Experimental results}
%%%%%%%%%%%%%%%%%%%%%%%%%%%%%%%%%%%%%%%%%%%%%%%%%%%%
\subsection{Dataset}
In the main experiment, the UNIPEN 1a isolated digit dataset (containing 11,078 samples in total) was used. Although digit samples show simple trajectories, they contain enough shape variations, such as straight line segments, corner points, curves with various curvatures, crossings, and closed loops (of `0'). 80\% was used for training and 20\% for testing. In the experiment of Section~\ref{sec:ex-alphabet}, the UNIPEN 1b isolated alphabet dataset (containing 15,661 samples) was also used. Each sample is resampled to be $T=50$ and represented as a sequence of three-dimensional vectors $(x_t, y_t, l_t)$ with length $T$ for the fixed-dimensional state representation of Section~\ref{sec:RL-for-HW}.
%------------------------------------------------
\subsection{Network structure\label{sec:modules}}
The network structures of the actor，the critic，and the discriminator are almost the same; two 1D-convolution layers with ReLU (kernel size=7 and stride=1) and then one fully-connected layer. The first convolutional layer is 128 channels and the second is 64. Their input is a 3-channel $T$-dimensional vector (i.e., a state). The output of the actor is $(x_t, y_t)\in [0,1]^2$. The critic and the discriminator output a scalar value. %------------------------------------------------
\subsection{Prediction-based model for comparative study}
As another promising handwriting generation model, we use
a prediction-based model. This model is a 1D-CNN with the same structure as the actor; therefore, its input is the partial trajectory with length $t$ and its output is $(x_{t+1}, y_{t+1})$.
The model, however, is trained to minimize the L2 norm
between its output and the real $(x_{t+1}, y_{t+1})$.
This means that the model tries to predict the next
pen-tip position based on the trajectory till $t$. To generate an entire trajectory by the trained model, the prediction is repeated till $T$ while concatenating the predicted pen-tip position with the past trajectory.
%------------------------------------------------
\subsection{Observation of the trained values}
Fig.~\ref{fig:learned-result}~(c) visualizes
the $Q(s_t, a_t)$-map, given by examining all possible $a_t$ for a fixed $s_t$.  For this figure, 15 test samples are randomly selected and their partial trajectories till $t=10, 20, 30$, and $40$ are used as $s_t$. Therefore, each
$Q(s_t, a_t)$-map shows the probability of the next pen-tip position for a real partial trajectory.\par
In most cases, the $Q$-map shows a unimodal distribution around the current pen-tip position $(x_t, y_t)$. A closer observation also reveals that the distribution is biased toward the moving direction. It should be noted again that no constraint is imposed on the range of the actions. The value maps, therefore, show that our model automatically acquires the basic characteristics of the handwriting trajectories.\par
Another important observation of the value maps is
anisotropic; namely, the value distribution is very different from Gaussian distribution, although it is still unimodal. This means that our model has more flexibility than typical parametric modeling schemes.
%=========================
\begin{figure}[t]
\centering
\includegraphics[width=\linewidth]{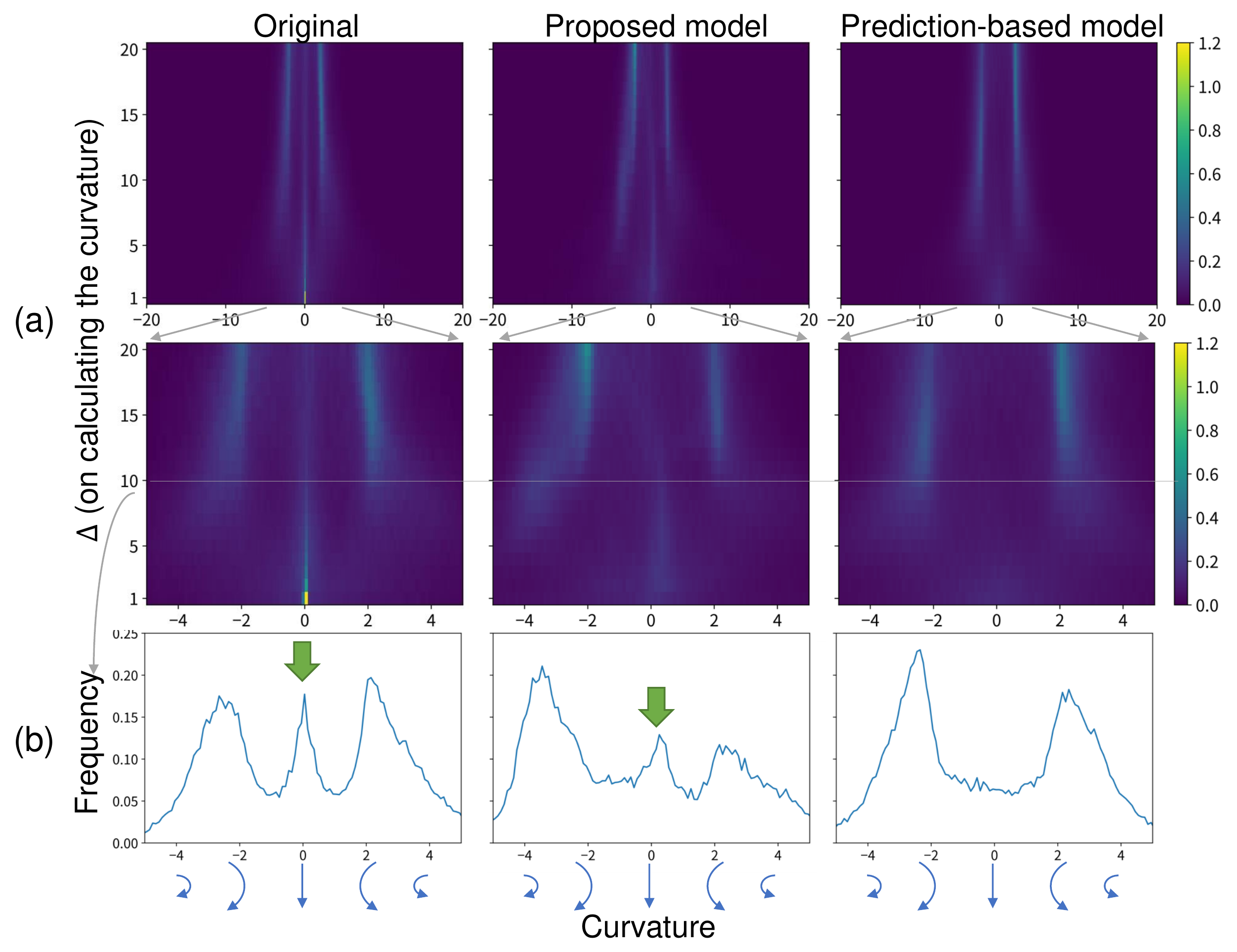}\\[-3mm]
\caption{(a)~Curvature distribution at different $\Delta$s. (b)~Curvature distribution of $\Delta=10$.}\vspace{-3mm}
\label{fig:delta_curvature}
\end{figure}
%=========================

%=========================
\begin{figure*}[t]
\centering
\includegraphics[width=\linewidth]{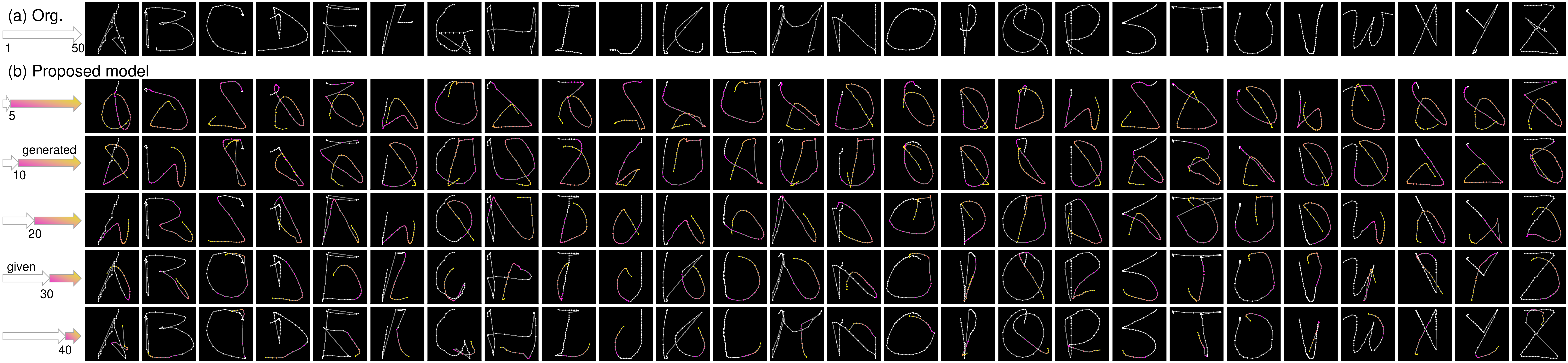}\\[-3mm]
\caption{Examples of the generated trajectories of English alphabets by our GAIL-based model.}
\label{fig:generated-alphabets}\vspace{-3mm}
\end{figure*}
%=========================
%------------------------------------------------
\subsection{Qualitative evaluation of generated trajectories}
Figs.~\ref{fig:generated}~(b) and (c) show the generated trajectories by our GAIL-based generation model and the prediction-based model, respectively.
Specifically, given beginning parts of the test samples of (a), the succeeding trajectories are generated till $t=T=50$. It should be noted that this is {\em not} a prediction task and therefore the generated parts need not be similar to the original. Instead, we expect that the generated parts have some handwriting-like shape.\par
Both models could generate handwriting-like trajectories. However, closer observation shows the superiority of the proposed model over the prediction-based model. For example, the proposed model generates trajectories that occupy the whole image region, whereas the prediction-based model sometimes cannot (as shown in the green-boxed cases). Controlling the whole size of the generated trajectories needs future planning throughout the generation steps. Therefore, this superiority comes from the future planning ability of the proposed model. \par
We can make a similar observation that the proposed model has less spurious parts around the end of the vertical trajectories, whereas the prediction-based model has more as shown in the blue-boxed cases. This indicates that our model controls the length of the trajectory by utilizing the future planning ability. \par
A more important superiority of the future planning ability is also found in the curvature of the generated trajectories.
The proposed model generates smoother trajectories with  lower curvatures, whereas the prediction-based model sometimes shows unnatural curves as shown in the orange-boxed cases. In fact, the generation of low-curvature trajectories requires careful future planning ability and, thus, our GAIL-based model is very appropriate for handwriting generation. In the next section, this superiority is confirmed by the quantitative evaluation with all test samples.
%------------------------------------------------
\subsection{Quantitative evaluation of generated trajectories}
A quantitative evaluation of the generated trajectories was conducted by using their curvature distribution. As noted before, our task is not to predict the future trajectory as accurately as possible; instead, we want to generate more handwriting-like trajectories. The curvature
distribution can be a reasonable choice to understand the general trend of the trajectory shapes. \par
A more important aspect is that the curvature can evaluate the trajectory shape in a multi-scale manner. The curvature is calculated as $1/r$ where $r$ is the radius of the circle passing three points $(x_{t-\Delta}, y_{t-\Delta}), (x_t, y_t), (x_{t+\Delta}, y_{t+\Delta})$ and different $\Delta$s give the curvatures in different scales. If $\Delta$ is large (small), the resulting curvature catches the global (local) structure of the trajectory. \par
Fig.~\ref{fig:delta_curvature}~(a) is the two-dimensional histograms showing the curvature distribution of the original trajectories, the generated trajectories by our model, and the generated trajectories by the prediction-based model. The beginning part ($t=1,\ldots,20$) of every test sample is fed to the model as $s_{20}$ and the remaining part ($t=21,\ldots, 50$) is generated. For each $t=21,\ldots, 50$ in the generated part, the curvature is calculated. The normalized histograms for $\Delta=1, \ldots, 20$ are visualized as the single two-dimensional histogram of Fig.~\ref{fig:delta_curvature}~(a).\par
Fig.~\ref{fig:delta_curvature}~(a) shows that our model can generate the trajectories having similar curvatures as the original. Especially, compared to the prediction-based model, our model can generate low-curvature trajectories more frequently as the original, as we expected from the smoother trajectories in Fig.~\ref{fig:generated}~(b).
Fig.~\ref{fig:delta_curvature}~(b) shows the normalized histogram at $\Delta=10$. As indicated by green arrows, a clear peak appears around the zero curvature for the original trajectory and the generated trajectory by ours and does not for the prediction-based model. \par
%
%------------------------------------------------
\subsection{Generation of Latin alphabets\label{sec:ex-alphabet}}
Fig.~\ref{fig:generated-alphabets} shows the example of generated trajectories for the alphabet samples which are randomly selected from UNIPEN isolated alphabet dataset. For this result, GAIL was trained with the alphabet dataset. Similarly to  Fig.~\ref{fig:generated}~(b), it is observed that the generated samples show handwriting-like curvatures while occupying the whole image region. From these results, it is possible to say that our model has enough capability to acquire the trends of arbitrary handwriting types.

%%%%%%%%%%%%%%%%%%%%%%%%%%%%%%%%%%%%%%%%%%%%%%%%%%%%
\section{Conclusion}
%%%%%%%%%%%%%%%%%%%%%%%%%%%%%%%%%%%%%%%%%%%%%%%%%%%%
In this paper, we developed a handwriting generation model using an RL framework, with the expectation that the future planning process in RL is useful to generate more realistic handwriting trajectories. Instead of using the typical RL framework, we employed an imitation learning algorithm called GAIL~\cite{GAIL}, where the reward function is trained along with other modules. This enhances the merit of using RL because we do not need to pre-define the reward function based on our intuition.
Experimental results show that the generated trajectories show similar curvatures to real handwriting trajectories. Considering the fact that generating trajectories with a specific (especially low) curvature requires an appropriate future (i.e., long-term) planning ability, these results indicate the usefulness of the proposed model.\par
Future work will focus on several points. First, it is necessary to derive a {\em general} handwriting model by mixing various handwriting types during training. Second, we need to utilize the trained model for discriminating handwriting trajectories from non-handwritings; for example, the comparative observation of their $Q$-maps will help us to understand what is the essential property of handwriting. Third, we can apply our model for handwriting-based medical diagnosis.

%%%%%%%%%%%%%%%%%%%%%%%%%%%%%%%%%%%%%%%%%%%%%%%%%%%%
\section*{Acknowledgment}
This work was partially supported by JSPS KAKENHI Grant Number JP17H06100.% \vskip 4mm
%%%%%%%%%%%%%%%%%%%%%%%%%%%%%%%%%%%%%%%%%%%%%%%%%%%%

%%%%%%%%%%%%%%%%%%%%%%%%%%%%%%%%%%%%%%%%%%%%%%%%%%%%
\bibliographystyle{IEEEtran}
\bibliography{icfhr}

% \begin{thebibliography}{00}
% \bibitem{kellogg} R. Kellogg, Analyzing Children's Art, National Press Books, 1969.
%書籍名だからダブルクオート不要
% \bibitem{doya}
% K. Doya, ``Complementary roles of basal ganglia and cerebellum in learningand motor control,'' Current Opinion in Neurobiology, vol.~10, no.~6, pp.~732–739, 2000.

% \bibitem{wada}
% Y.Wada and K. I. Sumita, ``A reinforcement learning scheme for acquisition of via-point representation of human motion,'' Proc. IEEE Int. Conf. Neural Networks, vol.~2, pp.~1109–1114, 2004.

% \bibitem{muhammad}
% U. R. Muhammad, Y. Yang, Y. Z. Song, T. Xiang, and T. M. Hospedales, ``Learning Deep Sketch Abstraction,” Proc. CVPR,  2018.

% \bibitem{chao}
% F. Chao, J. Lv, D. Zhou, L. Yang, C.-M. Lin, C. Shang, and C. Zhou, ``Generative adversarial nets in robotic Chinese calligraphy,''
% Proc. ICRA, pp. 1104–1110, 2018.

% \bibitem{wu}
% R. Wu, C. Zhou, F. Chao, L. Yang, C. M. Lin, and C. Shang, ``Integration of an actor-critic model and generative adversarial networks for a Chinese calligraphy robot,'' Neurocomputing, 2020.

% \end{thebibliography}

\end{document}